\def\year{2021}\relax
\documentclass[letterpaper]{article} 
\usepackage{amsmath,amssymb}
\usepackage{aaai21}  
\usepackage{times}  
\usepackage{helvet} 
\usepackage{courier}  
\usepackage[hyphens]{url}  
\usepackage{graphicx} 
\usepackage{natbib}  
\usepackage{caption} 
\title{MS-GWNN: Multi-Scale Graph Wavelet Neural Network \\ for Breast Cancer Diagnosis}
\author{Mo Zhang$^{1,2,3}$, Quanzheng Li$^{4}$
    \\
}
\affiliations{
    $^{1}$Center for Data Science, Peking University, Beijing 100871, China; $^{2}$Center for Data Science in Health and Medicine, Peking University, Beijing 100871, China; $^{3}$Laboratory for Biomedical Image Analysis, Beijing Institute of Big Data Research, Beijing 100871, China; $^{4}$Center for Advanced Medical Computing and Analysis, MGH/BWH Center for Clinical Data Science, Department of Radiology, Massachusetts General Hospital, Harvard Medical School, Boston, MA 02115, USA.

}

\begin{document}

\maketitle

\begin{abstract}
Breast cancer is one of the most common cancers in women worldwide, and early detection can significantly reduce the mortality rate of breast cancer. It is crucial to take multi-scale information of tissue structure into account in the detection of breast cancer. And thus, it is the key to design an accurate computer-aided detection (CAD) system to capture multi-scale contextual features in a cancerous tissue. In this work, we present a novel graph convolutional neural network for histopathological image classification of breast cancer. The new method, named multi-scale graph wavelet neural network (MS-GWNN), leverages the localization property of spectral graph wavelet to perform multi-scale analysis. By aggregating features at different scales, MS-GWNN can encode the multi-scale contextual interactions in the whole pathological slide. Experimental results on two public datasets demonstrate the superiority of the proposed method. Moreover, through ablation studies, we find that multi-scale analysis has a significant impact on the accuracy of cancer diagnosis.
\end{abstract}
\section{Introduction}
Breast cancer is the second leading cause of cancer-related death among women \cite{Rebecca2016Cancer}.  Effective treatment of breast cancer depends heavily on the accurate diagnosis at an early stage. In clinical practice, histopathological image analysis is the gold standard for detecting breast cancer, which is usually conducted manually by pathologists. However, this analysis is difficult even for skilled pathologists. According to the previous studies \cite{elmore2015diagnostic}, the average diagnostic concordance among different pathologists is only 75\%. Moreover, this diagnostic process is time-consuming due to the complexity of pathological images. Therefore, in order to improve accuracy and efficiency, it is necessary to develop computer-aided diagnosis systems (CADs) for cancer recognition.

In the past decades, many automatic algorithms based on digital pathology have been proposed for tumor classification. By exploiting handcrafted features, a variety of machine learning methods have been used, such as support vector machines (SVM) \cite{filipczuk2013computer}, multi-layer perceptron (MLP) \cite{george2013remote} and random forest (RF) \cite{nguyen2013random}. Currently, convolutional neural networks (CNNs) have achieved remarkable success in this field \cite{roy2019patch,yao2019parallel,alzubaidi2020optimizing}, benefiting from its advantage of extracting hierarchical features automatically. Typically, the whole pathological image is divided into small patches which are classified by a CNN, then these patch-wise predictions are integrated to obtain the final image-wise classification result. However, such patch-wise feature learning lacks the ability to capture global contextual information.

To overcome this limitation, some researchers have attempted to use graph convolutional networks (GCNs) for pathological image classification \cite{zhou2019cgc,anand2020histographs,wang2020weakly,adnan2020representation}. Most of these works follow the workflow below: Firstly, the pathological image is transformed into a graph representation, where the detected cancer cells serve as nodes and edges are formed in terms of spatial distance. Secondly, they extract cell-level features as the initial node embeddings. Thirdly, the cell-graph is fed into a GCN followed by a MLP to perform image-wise classification. In such setting, global features including spatial relations among cells are embeded in GCN. 

However, the current pipeline still faces two challenges. Firstly, only cellular interactions is insufficient to completely represent the pathological structure. In fact, the tissue distribution is hierarchical with many substructures, such as stromal, gland, tumor etc. To learn the intrinsic characteristic of cancerous tissue, it is necessary to aggregate multi-level structural information. For this reason, pathologists always need to analyze many images at different magnification levels to give an accurate cancer diagnosis. Secondly, this multi-stage workflow is tedious with too much workload. The performance of GCN relies heavily on the previous steps such as cell detection and feature extraction. Moreover, such staged framework lacks robustness to different datasets, as parameters need to be tuned at every step.

To tackle the aforementioned problems, we propose a novel framework named multi-scale graph wavelet neural network (MS-GWNN) for histopathological image classification. Graph wavelet neural network (GWNN) \cite{xu2019graph} replaces the graph Fourier transform in spectral GCN as graph wavelet transform. Further, GWNN has good localization property in node domain, making it more flexible to adjust the receptive fields of nodes (via the scaling parameter $s$). Based on GWNN, we present multi-scale graph wavelet neural network (MS-GWNN), which takes advantage of spectral graph wavelets to make multi-scale analysis. More specifically, after converting pathological images into graph representations, we use multiple GWNNs with different scaling parameters in parallel to obtain the multi-scale contextual information in graph topology. Then, all these features are aggregated to produce the final image-level (i.e. graph-level) classification prediction.

The main contributions are summarized as follows:
\begin{itemize}
    \item We propose a novel framework (MS-GWNN) for breast cancer diagnosis in the way of mapping pathological images into graph domain. By exploiting multi-scale graph wavelets, the proposed MS-GWNN can obtain multi-level tissue structural information, which is exactly what the pathology analysis needs. Although we apply MS-GWNN on a disease detection problem in this manuscript, the MS-GWNN is a general framework of image analysis that can be applied to many classifications tasks. 
    \item MS-GWNN can be trained in an end-to-end manner. Compared to the previous multi-stage workflow based on GCN, MS-GWNN simplifies the diagnostic process and enhances the robustness. To the best of our knowledge, MS-GWNN is the first end-to-end framework to apply GCN to pathological image classification.
    \item MS-GWNN is evaluated on two public breast cancer datasets (BACH and BreakHis), and it achieves an accuracy of 93.75\% and 99.67\% respectively. The results outperform the existing state-of-the-art methods, demonstrating the superiority of our proposed model. Through ablation studies, we verify that multi-scale structural features are crucial for characterizing cancers. 
\end{itemize}
\section{Related Work}
\subsection{Multi-Scale Feature Learning in Pathology Image} 
For cancer recognition, both local information about lesion appearance and global tissue organization are required. Based on this motivation, many attempts have been made to encode multi-scale information in pathology image, mainly including multi-scale features fusion \cite{bardou2018classification,shen2017multi,tokunaga2019adaptive} and the combination of CNN and recurrent neural networks (RNNs) \cite{zhou2018integrating,guo2018cnn,yan2020breast}. For instance, Shen et al. \cite{shen2017multi} presented a multi-crop pooling operation to enable a multi-level feature extraction, then these rich features are concatenated to form the final feature vector. In the work of \cite{tokunaga2019adaptive}, three parallel CNNs were used to process images at different magnifications, where the multi-branch features are incorporated in a weighted manner. In order to extract the contextual information in cancer tissue, Yan et al. \cite{yan2020breast} utilized an RNN to capture the spatial correlations between patch-wise features learned from a CNN. 

In summary, the above methods aim at learning hierarchical features in Euclidean space, where the notion of scale is typically related with Euclidean distance (spatial location). However, in this paper, we map pathological images into graph domain. The graph edges are formed according to the feature similarities between nodes, thus making it possible to explore the function of scale in non-Euclidean space.

\subsection{Spectral Graph Wavelet Transform} 
Spectral graph wavelet transform was originally proposed by Hammond et al. \cite{hammond2011wavelets}, which constructed wavelet transforms in the node domain of a finite weighted graph. They also proved that graph wavelets exhibit good localization property in the fine scale limit. In the work of \cite{tremblay2014graph}, Tremblay et al. applied graph wavelets to detect multi-scale communities in networks. They used the correlation between wavelets to access the similarity between nodes and proposed a clustering algorithm according to this similarity. Based on graph diffusion wavelets, Donnat et al. \cite{donnat2018learning} developed the GraphWave method to represent each node’s neighborhood via a low-dimensional embedding. This kind of node embedding pays more attention to the local topology of node. More recently, Xu et al. \cite{xu2019graph} proposed graph wavelet neural network (GWNN) to introduce the graph wavelet transform to spectral GCN, offering high spareness and good localization for graph convolution. However, their experiments were performed only at one scale, which didn't leverage the localization property of graph wavelets to perform multi-scale analysis.

\section{Method}
 In this section, we first introduce the basic concepts of spectral graph convolution, and further describe the spectral convolution based on graph wavelets. Then we discuss the architecture of graph wavelet neural network (GWNN). Finally, we present multi-scale graph wavelet neural network (MS-GWNN) for pathological image classification.
 
 \begin{figure*}[t]
	\centering
	\includegraphics[width=1.0\textwidth]{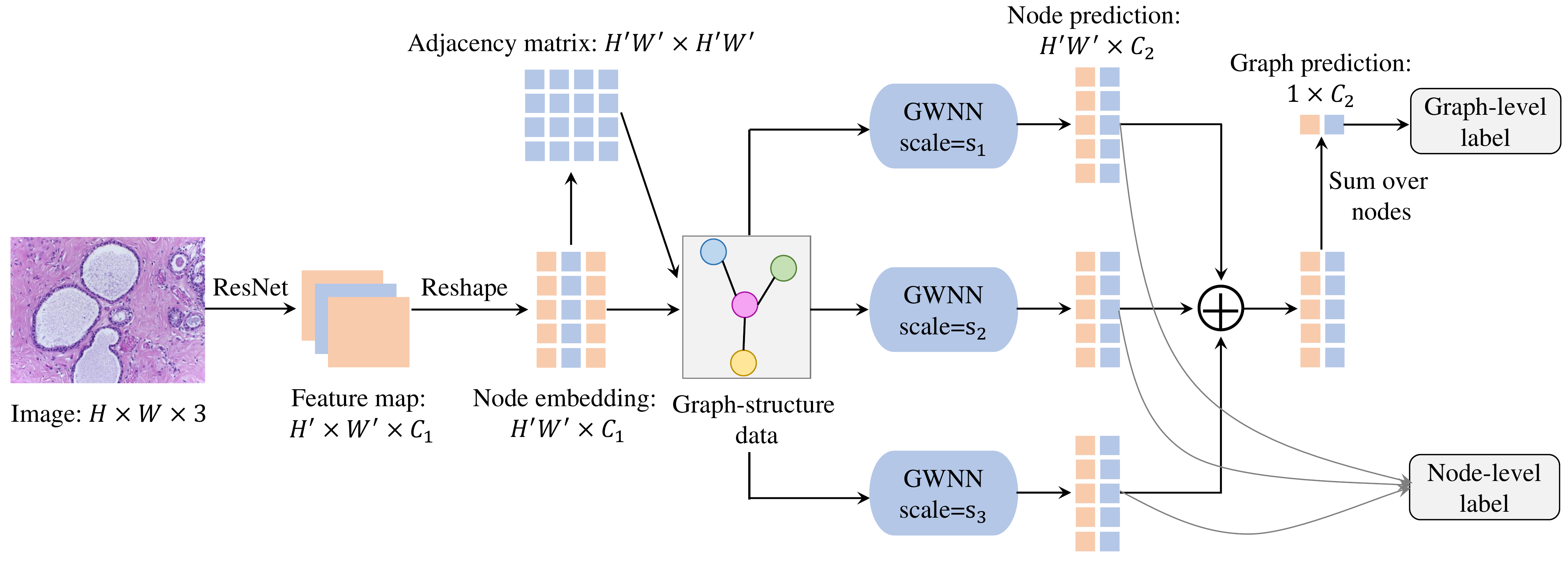}
	\caption{The architecture of MS-GWNN. Firstly, pathological images are transformed into graph-structure data. Then, node classification is performed via GWNN at different scales. Finally, the multi-level node embeddings are incorporated to yield the graph-level (image-level) classification result. $H^{'}W^{'}=N$ is the number of nodes. $C_1$ is the dimension of the initial node embeddings and $C_2$ is the number of cancer type. Both graph-level and node-level labels are used to train the model in an end-to-end way.}
	\label{fig:structure}
\end{figure*}
\subsection{Graph Representation} 
Let $G=\{V, E, \mathbf{H}\}$ be an undirected graph, where $V$ is the set of nodes with $|V|=N$ and $E$ is the set of edges. $\mathbf{H} \in \mathbb{R}^{N \times C}$ is the node embedding matrix. $\mathbf{A}$ is a symmetric adjacency matrix defining the graph topology. We denote the graph’s normalized Laplacian matrix as $\mathbf{L}=\mathbf{I_{N}}-\mathbf{D}^{-1 / 2} \mathbf{A} \mathbf{D}^{-1 / 2}$, where $\mathbf{I_{N}}$ is the identity matrix and $\mathbf{D}$ is a diagonal matrix with $D_{i, i}=\sum_{j} A_{i, j}$. Further, we denote $\mathbf{U}$ as the eigenvector decomposition of the normalized Laplacian matrix $\mathbf{L}=\mathbf{U} \mathbf{\Lambda} \mathbf{U}^{T}$, where $\mathbf{\Lambda}=\operatorname{Diag}\left(\lambda_{1}, \ldots, \lambda_{N}\right)$ is the diagonal matrix formed by the eigenvalues of $\mathbf{L}$.
\subsection{Spectral Graph Convolution} 
Spectral convolution was proposed by \cite{bruna2013spectral}, which defines the graph convolution operation in the Fourier domain. It can be represented as:
$$
\mathbf{g}_{\theta} \star \mathbf{x}=\mathbf{U} \mathbf{g}_{\theta}(\mathbf{\Lambda}) \mathbf{U}^{T} \mathbf{x}.
$$
$\mathbf{x} \in \mathbb{R}^{N}$ is the signal to be processed and $\mathbf{g}_{\theta}=\operatorname{Diag}(\boldsymbol{\theta})$ is the convolutional filter parameterized by $\boldsymbol{\theta} \in \mathbb{R}^{N}$. This definition of spectral graph convolution is not spatially localized in node domain. In other words, the feature aggregation of one node depends on all the nodes, not only its neighbourhood nodes. Moreover, it is computationally expensive to compute the eigendecomposition of Laplacian matrix. To improve computational efficiency, some researchers have attempted to use a truncated expansion based on Chebyshev polynomials to approximate the convolutional kernel \cite{hammond2011wavelets,defferrard2016convolutional,kipf2016semi}.
\subsection{Spectral Convolution Based on Graph Wavelets}
Spectral graph wavelet was presented in \cite{hammond2011wavelets}, which introduces a band-pass filter in the traditional graph Fourier domain mentioned above. We denote $\psi_{si}$ as the wavelet centered at node $i$ at the scale of $s$, thereby the graph wavelet basis is defined as:
$$
\mathbf{\Psi_{s}}=(\psi_{s1}, \psi_{s2}, \ldots, \psi_{sN})=\mathbf{U} \mathbf{H}_s \mathbf{U}^T.
$$
$\mathbf{H}_{s}=\operatorname{Diag}(h(s \lambda_{1}), \ldots, h(s \lambda_{N}))$ is a scaling matrix with $h(s \lambda_{i})=e^{\lambda_{i} s}$. Given a set of graph wavelets, the graph wavelet transform is defined as $\hat{\mathbf{x}}=\mathbf{\Psi}_{s}^{-1}\mathbf{x}$ and the inverse transform is $\mathbf{x}=\mathbf{\Psi}_{s}\hat{\mathbf{x}}$. $\mathbf{\Psi}_{s}^{-1}$ can be obtained by replacing $h(s \lambda_{i})$ in $\mathbf{\Psi}_{s}$ with $h(-s \lambda_{i})$. Further, in such setting, the graph convolution based on spectral wavelet bases reads as
$$
\mathbf{g}_{\theta} \star \mathbf{x}=\mathbf{\Psi}_{s} g_{\theta} \mathbf{\Psi}_{s}^{-1} \mathbf{x}.
$$
Similarly, $\mathbf{g}_{\theta}=\operatorname{Diag}(\boldsymbol{\theta})$ is the convolutional filter to be learned. As stated in the work of \cite{xu2019graph}, the graph convolution using wavelet transform has excellent localization property, making it outperforming the traditional spectral convolution in graph-related tasks like node classification. In addition, the scaling parameter $s$ controls the receptive fields of nodes in a continuous manner, different from the previous approach \cite{defferrard2016convolutional} using the discrete shortest path distance.
\subsection{Graph Wavelet Neural Network (GWNN)}
Based on spectral graph wavelet convolution, in this paper, we consider a three-layer graph wavelet neural network (GWNN) for node classification. The structure of the m-layer can be divided into two steps: 

feature transformation: $\mathbf{X}^{m^{\prime}}=\mathbf{X}^{m} \mathbf{W}^m$,

graph convolution: $\mathbf{X}^{m+1}=\sigma\left(\mathbf{\Psi}_{s} \mathbf{F}^{m} \mathbf{\Psi}_{s}^{-1} \mathbf{X}^{m^{\prime}}\right)$.\\
$\mathbf{X}^m \in \mathbb{R}^{N \times p}$ and $\mathbf{X}^{m+1} \in \mathbb{R}^{N \times q}$ are the input and output tensor respectively. $\mathbf{W}^m \in \mathbb{R}^{p \times q}$ is the matrix of filter parameters, $\mathbf{F}^m$ is the diagonal convolutional kernel, and $\sigma$ is the activation function. More specifically, our forward model can be described as:

first layer: $\mathbf{X}^{2}=relu \left(\mathbf{\Psi}_{s} \mathbf{F}^{1} \mathbf{\Psi}_{s}^{-1} \mathbf{X}^{1} \mathbf{W}^1 \right)$,

second layer: $\mathbf{X}^{3}=relu \left(\mathbf{\Psi}_{s} \mathbf{F}^{2} \mathbf{\Psi}_{s}^{-1} \mathbf{X}^{2} \mathbf{W}^2 \right)$,

third layer: $\mathbf{X}^{4}=softmax \left(\mathbf{\Psi}_{s} \mathbf{F}^{3} \mathbf{\Psi}_{s}^{-1} \mathbf{X}^{3} \mathbf{W}^3 \right)$.
\subsection{Multi-Scale Graph Wavelet Neural Network (MS-GWNN)}
In this section, we present a new framework called multi-scale graph wavelet neural network (MS-GWNN), for the task of pathological image classification. The architecture of MS-GWNN is shown in Figure \ref{fig:structure}, which consists of three parts: graph construction, node classification based on GWNN and graph classification using feature aggregation.

\textbf{Graph Construction.}
In this work, we transform pathological images into graph representations. Nodes are nonoverlapping image patches and edges are generated in terms of the intrinsic relationships between these patches. Firstly, we use the modified ResNet-50 (we remove block3,4 and use average-pooling for downsampling) to learn discriminative features. In this way, we can obtain a series of feature maps ($H^{'} \times W^{'} \times C_1$), where each pixel corresponds to a $r \times r$ ($r$ is the downsampling rate) square patch in the raw image. Therefore, the feature vector of one pixel can be regarded as the node embedding of the corresponding patch (node).

Secondly, we make use of the similarity between node embeddings to form edges, which is defined by dot product:
$$
f\left(\mathbf{x}_{i}, \mathbf{x}_{j}\right)= \theta\left(\mathbf{x}_{i}\right)^{\mathrm{T}} \phi\left(\mathbf{x}_{j}\right).
$$
$\mathbf{x}_i,\mathbf{x}_j$ are the node embeddings of node $i$ and $j$, while $\theta(\mathbf{x}),\phi(\mathbf{x})$ are two transformation functions implemented via $1 \times 1$ convolutions. The formula to form graph edges is as follow:
$$
e_{ij}=\left\{\begin{array}{ll}
{1,} & {\text {if} \ i=j \ \text{or} \ f\left(\mathbf{x}_{i}, \mathbf{x}_{j}\right)\geq Q_{\alpha}(f)} \\
{0,} & {\text {otherwise}}
\end{array}\right.
$$
$Q_{\alpha}(f)$ is the $\alpha$-th percentile ($0< \alpha \leq 100$) of $\{f\left(\mathbf{x}_{i}, \mathbf{x}_{j}\right)|i,j=1,2,\ldots,N\}$. $N$ is the number of nodes equivalent to $H^{'}W^{'}$ here.

\textbf{Node Classification via GWNN.}
Given the constructed graph-structure data, we utilize multiple GWNNs with different scaling parameters $s$ in parallel to conduct node classification. Each branch is supervised by the node-level label via the cross entropy loss function. Different parameters $s$ mean different receptive fields, enabling the model to extract multi-scale contextual information.

\textbf{Feature Aggregation.}
After the process of node classification, each branch can obtain a node prediction probability map with the size of $H^{'}W^{'}\times C_2$ ($C_2$ is the number of cancer type). In this work, we sum these probability maps to aggregate multi-scale structure representations. To perform graph-level (i.e. image-level) classification, we sum the node class probabilities to yield the graph-level prediction probability. Then, this graph-level prediction is passed to a softmax layer, which is supervised by the graph-level label.

In this framework, the final loss is formed by the node-level loss and graph-level loss together as follow:
\begin{equation}
\mathcal{L}_{final}=\lambda \cdot \mathcal{L}_{node}+\mathcal{L}_{graph},
\end{equation}
where $\lambda$ is a balancing parameter. Note that both $\mathcal{L}_{node}$ and $\mathcal{L}_{graph}$ are implemented via the classic cross entropy loss function, and $\mathcal{L}_{node}$ is the sum of the node-level losses from three branches. By minimizing the final loss, the error can be easily back-propagated through the whole MS-GWNN system in an end-to-end way. 
\section{Experiments and Results}
\subsection{Datasets}
We validate the proposed MS-GWNN on two public datasets: ICIAR 2018
breast cancer histology (BACH) grand challenge dataset \cite{aresta2019bach} and BreakHis dataset \cite{spanhol2015dataset}. The BACH dataset consists of 400 histopathological images with size of $2048 \times 1536 \times 3$, which aims to predict breast cancer type as 1) normal, 2) benign, 3) in situ carcinoma, and 4) invasive carcinoma. We randomly select 320 samples for training and the rest 80 images for testing. In BreakHis dataset, there are 7909 samples classified as either benign or malignant. The $700 \times 460$ images are collected at different magnification factors ($40\times$, $100\times$, $200\times$, $400\times$). Experiments are conducted on the samples obtained at $40\times$ magnification (1995 images in total). In line with the previous works \cite{bardou2018classification,gandomkar2018mudern,alom2019breast}, the entire dataset is randomly divided into a training set with 70\% samples and a testing set with 30\% data. In this work, all images on BACH (BreakHis) dataset are resized to $512\times 384$ ($350\times 230$) to reduce the memory workload of GPU. For image preprocessing, we use the classic H\&E color normalization approach described in \cite{vahadane2015structure}. To avoid overfitting, extensive data augmentation is performed including flip, rotation, translation, shear and linear contrast normalization. Moreover, the performance of model is evaluated according to the average accuracy at the image level.
\subsection{Implementations}
To implement the proposed model, we use the Tensorflow platform on an NVIDIA GeForce GTX 1080 Ti GPU. Each GWNN model consists of three graph convolutional layers with the feature dimension of 256, 128 and 4 (2) respectively. The overall downsampling rate of the modified ResNet-50 is $1/16$. Moreover, we use the Adam optimizer to train the model, where the initial learning rate is $10^{-3}$ and $\beta_1 = 0.9, \beta_2 = 0.99$. All models are trained for 30000 epochs with a batch size of 16. The hyperparameters $\alpha,\lambda$ are set as 99 and 1 respectively.
\subsection{Computational Complexity}
As the eigendecomposition of Laplacian matrix requires much computational cost, we utilize a fast algorithm to approximate the graph wavelets. Hammond et al. \cite{hammond2011wavelets} suggested that $\mathbf{\Psi}_{s}$ and $\mathbf{\Psi}_{s}^{-1}$ can be efficiently approximated by a truncated expansion according to the Chebyshev polynomials with low orders. In this way, the computational complexity is $O(k\times|E|),$ where $|E|$ is the number of graph edges and $k$ is the degree of Chebyshev polynomials. In this work, we set $k=2$. 
\subsection{Results and Comparisons}
\begin{figure}[t]
	\centering
	\includegraphics[width=0.48\textwidth]{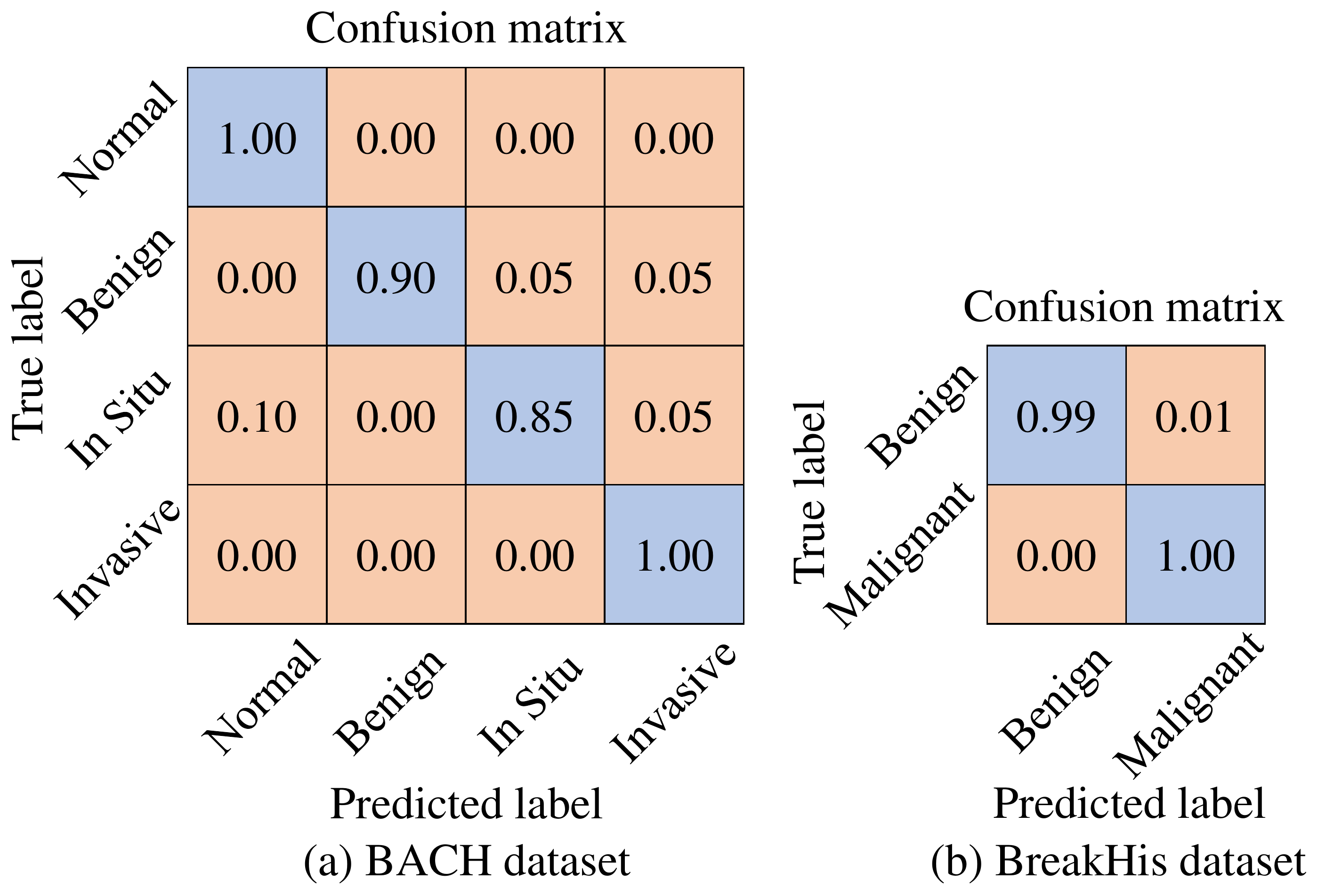}
	\caption{Normalized confusion matrix results of MS-GWNN on two datasets.}
	\label{fig:CM}
\end{figure}
\begin{table}[t]
	\begin{center}
		\caption{Comparisons with state-of-the-art methods on BACH dataset. }
		\label{table:bach}
		\scalebox{1.0}{
			\begin{tabular}{c|c}
				\hline
				Model & Accuracy\\ 
				\hline
			    \cite{golatkar2018classification} &  85.00\% \\
			    \hline
			    \cite{mahbod2018breast} &  88.50\% \\
				\hline
			    \cite{roy2019patch} & 90.00\% \\
			    \hline
			    \cite{meng2019multi} & 91.00\% \\
			    \hline
			    \cite{yao2019parallel} & 92.00\% \\
			    \hline
			   \cite{wang2018classification} & 92.00\% \\
			    \hline
			    \cite{kassani2019breast} & 92.50\% \\
			    \hline
			     \textbf{Proposed MS-GWNN}  &  \textbf{93.75\%} \\
			    
				\hline
		\end{tabular}}
	\end{center}
\end{table}
\begin{table}[t]
	\begin{center}
		\caption{Comparisons with state-of-the-art methods on BreakHis dataset. }
		\label{table:breakhis}
		\scalebox{0.9}{
			\begin{tabular}{c|c}
				\hline
				Model & Accuracy\\ 
				\hline
			    \cite{song2017adapting} &  87.00\% \\
			    	\hline
			     \cite{nahid2019histopathological}& 95.00\%  \\
			     	\hline
			     \cite{han2017breast}&  95.80\% \\
			     	\hline
			     \cite{veeling2018rotation}&   96.10\% \\
			     	\hline
			     \cite{kausar2019hwdcnn}&   97.02\%\\
			     	\hline
			     \cite{alom2019breast}&   97.95\%\\
			     	\hline
			     \cite{bardou2018classification}&  98.33\% \\
			     	\hline
			     \cite{gandomkar2018mudern}&  98.60\% \\
			    \hline
			     \textbf{Proposed MS-GWNN} &  \textbf{99.67\%} \\
			    
				\hline
		\end{tabular}}
	\end{center}
\end{table}
\begin{figure*}[t]
	\centering
	\includegraphics[width=0.9\textwidth]{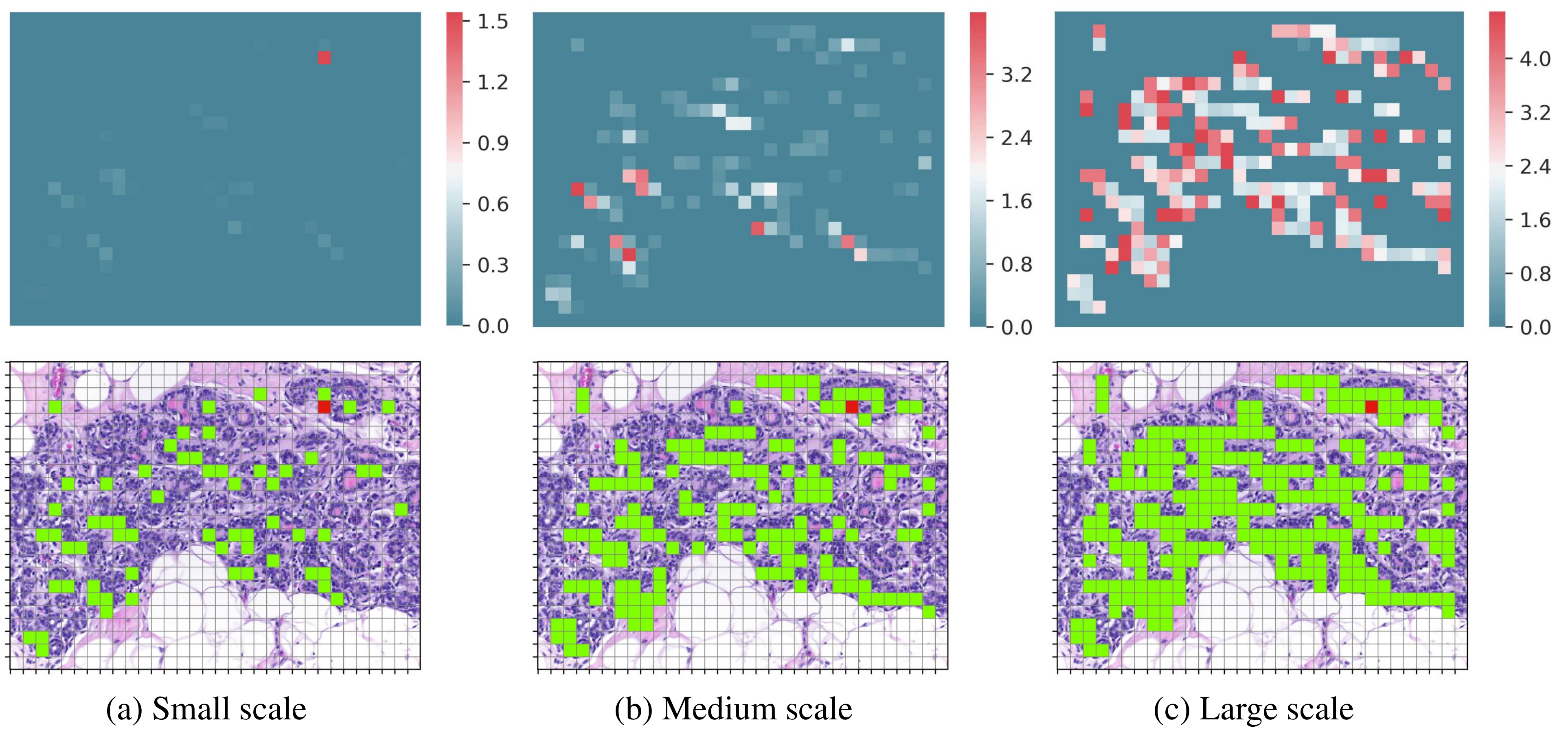}
	\caption{Visualization of the wavelet bases at different scales. The small, medium and large scale correspond to $s=1,3,5$ respectively. Top row: multi-scale graph wavelets $\Psi_{si}$ centered at node $i$. Bottom row: the receptive fields corresponding to the above wavelets. Each square denotes a node (patch) in the graph (image). The red point is the center node $i$ and the other green points are the neighborhood of node $i$. As the scaling parameter $s$ gets larger, the receptive field of node $i$ becomes wider accordingly.}
	\label{fig:multiscale}
\end{figure*}

As for breast cancer classification, the new MS-GWNN obtains an accuracy of 93.75\% and 99.67\% on BACH and BreakHis dataset respectively. The normalized confusion matrices are shown in Figure \ref{fig:CM}. On BACH dataset, both normal and invasive categories yield a high accuracy of 100\%, while the accuracy of predicting in situ type is relatively low. This could be due to the similar structures appeared in inter-class images. In addition, we compare the results with existing state-of-the-art models as listed in Table \ref{table:bach},\ref{table:breakhis}. Obviously, MS-GWNN outperforms the other methods on both datasets, especially on BreakHis dataset where the error rate is only 0.33\%. The results demonstrate the strong capacity of the proposed MS-GWNN model to tackle pathological image classification.
\section{Ablation Studies}
\begin{table}[t]
	\begin{center}
		\caption{Ablation study of the multi-scale feature learning on BACH dataset. We set $\lambda$ as 1 in all the experiments. MS-GWNN-1 ($s=0.5$) denotes the MS-GWNN model with one branch ($s=0.5$).}
		\label{table:ablation}
		\scalebox{1.0}{
			\begin{tabular}{c|c}
				\hline
				Model & Accuracy\\ 
				\hline
				GCN      & 88.75\%     \\
				\hline
			    MS-GWNN-1 (s=0.5) & 88.75\% \\
			    MS-GWNN-1 (s=1.0) & 88.75\% \\
			    MS-GWNN-1 (s=1.5) &  87.50\% \\
			    \hline
			    MS-GWNN-2 (s=0.5,1.0) & 91.25\% \\
			    MS-GWNN-2 (s=0.5,1.5) &  92.50\% \\
			    MS-GWNN-2 (s=1.0,1.5) &  91.25\% \\
			    \hline
			    \textbf{MS-GWNN-3 (s=0.5,1.0,1.5)} &  \textbf{93.75\%} \\
			    
				\hline
		\end{tabular}}
	\end{center}
\end{table}
To verify the effectiveness of MS-GWNN, we perform two kinds of ablation tests on BACH dataset. One is to investigate the effect of multi-scale feature extraction, and the other is about the selection of the balancing parameter $\lambda$ in loss.
\begin{figure*}[t]
	\centering
	\includegraphics[width=1.0\textwidth]{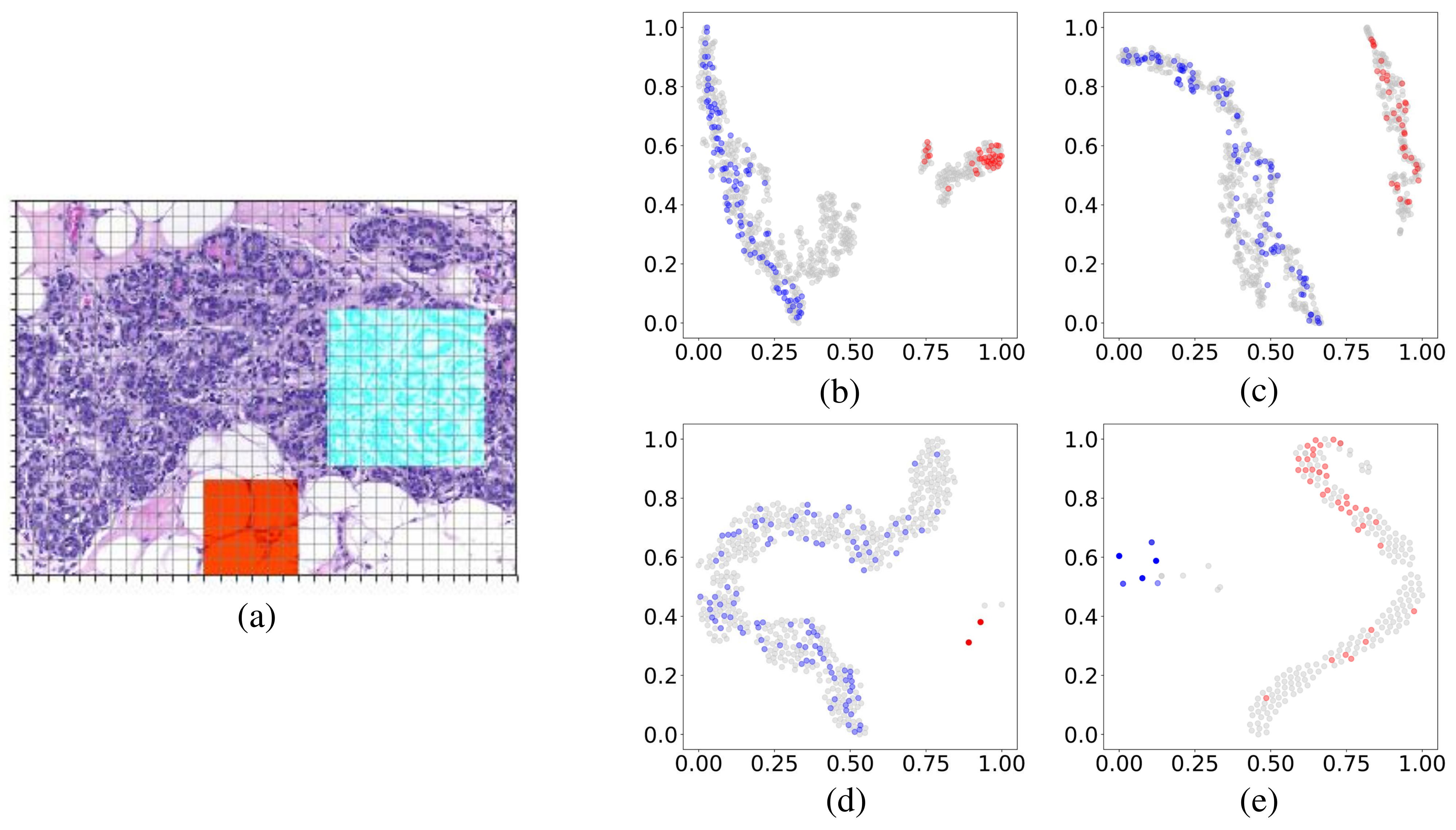}
	\caption{Representations of different node embeddings using t-SNE. (a) Raw image. Each square denotes a node in the graph. (b) t-SNE representations of the initial node embeddings (learned from ResNet). (c)(d)(e) t-SNE representations of the node embeddings produced by GWNN at the small (medium,large) scale. The red (blue) points in (b)(c)(d)(e) correspond to the red (blue) squares in the raw image (a). The node embeddings produced by GWNN at different scales exploit multi-level tissue structural features, providing richer information for cancer diagnosis. }
	\label{fig:tSNE}
\end{figure*}
\subsection{The Function of Multi-Scale Feature Learning}
In this section, we aim at analyzing the effect of the multi-scale features fusion. As shown in Table \ref{table:ablation}, as aggregating features at more scales, the model performance becomes better accordingly. The specific scaling parameter $s$ is given in brackets. When using only one-level features ($s=0.5$), MS-GWNN obtains an accuracy of 88.75\%. After introducing the features at another scale ($s=1.0$), the accuracy increases to 91.25\%. Further, MS-GWNN with three branches ($s=0.5,1.0,1.5$) can achieve a 93.75\% accuracy. This comparison confirms that extracting multi-scale context information is very important for pathological image analysis, which is consist with the experience of pathologists.

In addition, we also perform an experiment which replaces the multi-branch GWNNs in MS-GWNN as the traditional graph convolutional network (GCN) \cite{kipf2016semi}. As shown in Table \ref{table:ablation}, GCN achieves a similar result to MS-GWNN with one branch. However, due to the high spareness of graph wavelets, GWNN is much more computationally efficient than GCN, which has also been discussed in the previous work \cite{xu2019graph}.

\textbf{Visualization of the wavelet bases at different scales.}
To further investigate the working mechanism of MS-GWNN, we show the graph wavelet bases at different scales in the top row of Figure \ref{fig:multiscale}. Each small square represents a node in the constructed graph, and the node-to-patch correspondence relation is drawn in the raw image as shown in the second row of Figure \ref{fig:multiscale}. Specifically, the red square denotes the node $i$ which the wavelet $\Psi_{si}$ centered at, and the green squares denote the neighbourhood of node $i$ whose values are greater than 0 in the wavelet $\Psi_{si}$. As the scale gets larger, the scope of neighborhood becomes wider accordingly, meaning that the receptive field of node $i$ is gradually expanding. At the small scale ($s=1$), only a minority of nodes contribute to the embedding updating of node $i$. However, at the large scale ($s=5$), the neighborhood nodes increase a lot which nearly spread on the entire tissue. In such setting, information can be propagated among the tissue structures at different levels, enabling MS-GWNN to acquire multi-scale contextual features.

\textbf{Visualization of the learned node embeddings.}
In order to find the intrinsic characteristics of the node embeddings, we plot the 2D t-SNE projections of node embeddings in Figure \ref{fig:tSNE}. T-distributed stochastic neighbor embedding (t-SNE) can convert high-dimensional data to a low-dimensional (2-dimensional here) representation while maintaining the distance between objects. Then, the extracted 2D vectors are normalized to $[0,1]$ and are plotted in a scatter diagram. In Figure \ref{fig:tSNE}, (a) is the raw image where the squares correspond to the patches (nodes) in the image (graph). (b) shows the t-SNE representation of the initial node embeddings which are learned from the modified ResNet. (c)(d)(e) are the t-SNE representations of the node embeddings yielded by GWNN with different scales. It should be noted that the red and blue points in the scatter diagrams correspond to the squares (nodes) with the same color in (a). Obviously, in the pathological image (a), the red and blue squares belong to different tissues. Similarly, in (b)(c)(d)(e), the red and blue points are located in the different clusters. This phenomenon demonstrate that the intrinsic characteristics of nodes have been encoded in the embeddings.

On the other hand, the distribution of points reflects the inherent relationship between nodes. At the medium scale ($s=3$), the red points are very close to each other. However, at the large scale (s=5), the red points have a much wider distribution. This contrast prove that the node embeddings produced by GWNN at different scales encode different structural information. Compared to using the CNN features only, MS-GWNN can provide richer node embeddings representing multi-level tissue structures. In clinical diagnosis, the tissue structural information is a critical factor considered by pathologists.

\textbf{The Selection of the Scaling Parameter $s$.}
\begin{table}[t]
	\begin{center}
		\caption{Ablation study of the parameter $\lambda$ on BACH dataset. }
		\label{table:ablation2}
		\scalebox{1.0}{
			\begin{tabular}{c|c}
				\hline
				Model & Accuracy\\ 
				\hline
			    MS-GWNN-3 ($\lambda=0.01$) & 90.00\%  \\
			    \hline
			    MS-GWNN-3 ($\lambda=0.1$) & 92.50\%  \\
			    \hline
			    \textbf{MS-GWNN-3 ($\lambda=1$)} &  \textbf{93.75\%} \\
			    \hline
			    MS-GWNN-3 ($\lambda=10$) & 91.25\%  \\
			    \hline
			    MS-GWNN-3 ($\lambda=100$) & 88.75\%  \\
			    
				\hline
		\end{tabular}}
	\end{center}
\end{table}
As mentioned before, we can adjust the receptive field by varying the parameter $s$. Hence how to select the appropriate scale $s$ is a key factor for GWNN.  According to our experience of parameter tuning, a rule of thumb is to select the scaling parameter $s$ in $[0,2]$. if the scale is much greater than 2, the model performance will decrease. There are methods to guide the selection of $s$ in graph wavelet analysis. Donnat et al. \cite{donnat2018learning} designed a technique for selecting a set of applicable scales. In their strategy, the maximum and minimum scale are set to $-\log (\eta) / \sqrt{\lambda_{2} \lambda_{N}}$ and $-\log (\gamma) / \sqrt{\lambda_{2} \lambda_{N}}$ respectively. Then the appropriate scale can be selected in $[s_{min},s_{max}]$. $\lambda_{N}$ is the maximum eigenvalue of Laplacian matrix $\mathbf{L}$, and $\lambda_{2}$ is the minimum nonzero eigenvalue. They suggest setting $\eta=$ 0.85 and $y=0.95$. However, our experiments show this strategy doesn't work for graph wavelet neural network. We will further investigate how to design an elegant, theoretically guided rule to select the scaling parameter $s$. 

\subsection{The Selection of Parameter $\lambda$}
Table \ref{table:ablation2} shows the numerical results of MS-GWNNs with different parameter $\lambda$. The number ``3" in ``MS-GWNN-3" means that the MS-GWNN model utilizes three branches to learn multi-scale structural information. In more details, the three scaling factors in this experiment are set as $0.5,1.0,1.5$. It can be observed that the parameter $\lambda$ is an important factor influencing the performance of MS-GWNN. Further, the best option is $\lambda=1$, suggesting that 
both graph-level and node-level supervisions are equally crucial to identify cancer types.
\section{Conclusion}
In this work, we propose multi-scale graph wavelet neural network (MS-GWNN) for histopathological image classification. The MS-GWNN leverages the localization property of graph wavelets to perform multi-scale analysis with different scaling parameter $s$. For the task of breast cancer diagnosis, MS-GWNN outperforms the state-of-the-art approaches, mainly resulting from its powerful ability to integrate multi-scale contextual interactions. Through ablation studies, we prove the importance of exploiting multi-scale features. More broadly, the MS-GWNN model provides a novel solution to extract multi-scale structural information using graph wavelets, which can also be applied to other tasks.
\section{Ethics Statement}
Across the world, breast cancer has frequent morbidity and high mortality among women. Early detection is the key to increase the survival rate. Our proposed MS-GWNN has great performance on breast cancer diagnosis with high efficiency, which is valuable for the clinical applications. Further, it opens up a new direction towards better multi-scale representations of pathological images in graph domain.
\bibliography{gwnn}

\begin{thebibliography}{40}
\providecommand{\natexlab}[1]{#1}
\providecommand{\url}[1]{\texttt{#1}}
\providecommand{\urlprefix}{URL }
\expandafter\ifx\csname urlstyle\endcsname\relax
  \providecommand{\doi}[1]{doi:\discretionary{}{}{}#1}\else
  \providecommand{\doi}{doi:\discretionary{}{}{}\begingroup
  \urlstyle{rm}\Url}\fi

\bibitem[{Adnan, Kalra, and Tizhoosh(2020)}]{adnan2020representation}
Adnan, M.; Kalra, S.; and Tizhoosh, H.~R. 2020.
\newblock Representation Learning of Histopathology Images using Graph Neural
  Networks.
\newblock In \emph{Proceedings of the IEEE/CVF Conference on Computer Vision
  and Pattern Recognition Workshops}, 988--989.

\bibitem[{Alom et~al.(2019)Alom, Yakopcic, Nasrin, Taha, and
  Asari}]{alom2019breast}
Alom, M.~Z.; Yakopcic, C.; Nasrin, M.~S.; Taha, T.~M.; and Asari, V.~K. 2019.
\newblock Breast cancer classification from histopathological images with
  inception recurrent residual convolutional neural network.
\newblock \emph{Journal of digital imaging} 32(4): 605--617.

\bibitem[{Alzubaidi et~al.(2020)Alzubaidi, Al-Shamma, Fadhel, Farhan, Zhang,
  and Duan}]{alzubaidi2020optimizing}
Alzubaidi, L.; Al-Shamma, O.; Fadhel, M.~A.; Farhan, L.; Zhang, J.; and Duan,
  Y. 2020.
\newblock Optimizing the Performance of Breast Cancer Classification by
  Employing the Same Domain Transfer Learning from Hybrid Deep Convolutional
  Neural Network Model.
\newblock \emph{Electronics} 9(3): 445.

\bibitem[{Anand, Gadiya, and Sethi(2020)}]{anand2020histographs}
Anand, D.; Gadiya, S.; and Sethi, A. 2020.
\newblock Histographs: graphs in histopathology.
\newblock In \emph{Medical Imaging 2020: Digital Pathology}, volume 11320,
  113200O. International Society for Optics and Photonics.

\bibitem[{Aresta et~al.(2019)Aresta, Ara{\'u}jo, Kwok, Chennamsetty, Safwan,
  Alex, Marami, Prastawa, Chan, Donovan et~al.}]{aresta2019bach}
Aresta, G.; Ara{\'u}jo, T.; Kwok, S.; Chennamsetty, S.~S.; Safwan, M.; Alex,
  V.; Marami, B.; Prastawa, M.; Chan, M.; Donovan, M.; et~al. 2019.
\newblock Bach: Grand challenge on breast cancer histology images.
\newblock \emph{Medical image analysis} 56: 122--139.

\bibitem[{Bardou, Zhang, and Ahmad(2018)}]{bardou2018classification}
Bardou, D.; Zhang, K.; and Ahmad, S.~M. 2018.
\newblock Classification of breast cancer based on histology images using
  convolutional neural networks.
\newblock \emph{IEEE Access} 6: 24680--24693.

\bibitem[{Bruna et~al.(2013)Bruna, Zaremba, Szlam, and
  LeCun}]{bruna2013spectral}
Bruna, J.; Zaremba, W.; Szlam, A.; and LeCun, Y. 2013.
\newblock Spectral networks and locally connected networks on graphs.
\newblock \emph{arXiv preprint arXiv:1312.6203} .

\bibitem[{Defferrard, Bresson, and
  Vandergheynst(2016)}]{defferrard2016convolutional}
Defferrard, M.; Bresson, X.; and Vandergheynst, P. 2016.
\newblock Convolutional neural networks on graphs with fast localized spectral
  filtering.
\newblock In \emph{Advances in neural information processing systems},
  3844--3852.

\bibitem[{Donnat et~al.(2018)Donnat, Zitnik, Hallac, and
  Leskovec}]{donnat2018learning}
Donnat, C.; Zitnik, M.; Hallac, D.; and Leskovec, J. 2018.
\newblock Learning structural node embeddings via diffusion wavelets.
\newblock In \emph{Proceedings of the 24th ACM SIGKDD International Conference
  on Knowledge Discovery \& Data Mining}, 1320--1329.

\bibitem[{Elmore et~al.(2015)Elmore, Longton, Carney, Geller, Onega, Tosteson,
  Nelson, Pepe, Allison, Schnitt et~al.}]{elmore2015diagnostic}
Elmore, J.~G.; Longton, G.~M.; Carney, P.~A.; Geller, B.~M.; Onega, T.;
  Tosteson, A.~N.; Nelson, H.~D.; Pepe, M.~S.; Allison, K.~H.; Schnitt, S.~J.;
  et~al. 2015.
\newblock Diagnostic concordance among pathologists interpreting breast biopsy
  specimens.
\newblock \emph{Jama} 313(11): 1122--1132.

\bibitem[{Filipczuk et~al.(2013)Filipczuk, Fevens, Krzy{\.z}ak, and
  Monczak}]{filipczuk2013computer}
Filipczuk, P.; Fevens, T.; Krzy{\.z}ak, A.; and Monczak, R. 2013.
\newblock Computer-aided breast cancer diagnosis based on the analysis of
  cytological images of fine needle biopsies.
\newblock \emph{IEEE transactions on medical imaging} 32(12): 2169--2178.

\bibitem[{Gandomkar, Brennan, and Mello-Thoms(2018)}]{gandomkar2018mudern}
Gandomkar, Z.; Brennan, P.~C.; and Mello-Thoms, C. 2018.
\newblock MuDeRN: Multi-category classification of breast histopathological
  image using deep residual networks.
\newblock \emph{Artificial intelligence in medicine} 88: 14--24.

\bibitem[{George et~al.(2013)George, Zayed, Roushdy, and
  Elbagoury}]{george2013remote}
George, Y.~M.; Zayed, H.~H.; Roushdy, M.~I.; and Elbagoury, B.~M. 2013.
\newblock Remote computer-aided breast cancer detection and diagnosis system
  based on cytological images.
\newblock \emph{IEEE Systems Journal} 8(3): 949--964.

\bibitem[{Golatkar, Anand, and Sethi(2018)}]{golatkar2018classification}
Golatkar, A.; Anand, D.; and Sethi, A. 2018.
\newblock Classification of breast cancer histology using deep learning.
\newblock In \emph{International Conference Image Analysis and Recognition},
  837--844. Springer.

\bibitem[{Guo et~al.(2018)Guo, Liu, Bakker, Guo, and Lew}]{guo2018cnn}
Guo, Y.; Liu, Y.; Bakker, E.~M.; Guo, Y.; and Lew, M.~S. 2018.
\newblock CNN-RNN: A large-scale hierarchical image classification framework.
\newblock \emph{Multimedia Tools and Applications} 77(8): 10251--10271.

\bibitem[{Hammond, Vandergheynst, and Gribonval(2011)}]{hammond2011wavelets}
Hammond, D.~K.; Vandergheynst, P.; and Gribonval, R. 2011.
\newblock Wavelets on graphs via spectral graph theory.
\newblock \emph{Applied and Computational Harmonic Analysis} 30(2): 129--150.

\bibitem[{Han et~al.(2017)Han, Wei, Zheng, Yin, Li, and Li}]{han2017breast}
Han, Z.; Wei, B.; Zheng, Y.; Yin, Y.; Li, K.; and Li, S. 2017.
\newblock Breast cancer multi-classification from histopathological images with
  structured deep learning model.
\newblock \emph{Scientific reports} 7(1): 1--10.

\bibitem[{Kassani et~al.(2019)Kassani, Kassani, Wesolowski, Schneider, and
  Deters}]{kassani2019breast}
Kassani, S.~H.; Kassani, P.~H.; Wesolowski, M.~J.; Schneider, K.~A.; and
  Deters, R. 2019.
\newblock Breast cancer diagnosis with transfer learning and global pooling.
\newblock \emph{arXiv preprint arXiv:1909.11839} .

\bibitem[{Kausar et~al.(2019)Kausar, Wang, Idrees, and Lu}]{kausar2019hwdcnn}
Kausar, T.; Wang, M.; Idrees, M.; and Lu, Y. 2019.
\newblock HWDCNN: Multi-class recognition in breast histopathology with Haar
  wavelet decomposed image based convolution neural network.
\newblock \emph{Biocybernetics and Biomedical Engineering} 39(4): 967--982.

\bibitem[{Kipf and Welling(2016)}]{kipf2016semi}
Kipf, T.~N.; and Welling, M. 2016.
\newblock Semi-supervised classification with graph convolutional networks.
\newblock \emph{arXiv preprint arXiv:1609.02907} .

\bibitem[{Mahbod et~al.(2018)Mahbod, Ellinger, Ecker, Smedby, and
  Wang}]{mahbod2018breast}
Mahbod, A.; Ellinger, I.; Ecker, R.; Smedby, {\"O}.; and Wang, C. 2018.
\newblock Breast cancer histological image classification using fine-tuned deep
  network fusion.
\newblock In \emph{International Conference Image Analysis and Recognition},
  754--762. Springer.

\bibitem[{Meng, Zhao, and Su(2019)}]{meng2019multi}
Meng, Z.; Zhao, Z.; and Su, F. 2019.
\newblock Multi-classification of Breast Cancer Histology Images by Using
  Gravitation Loss.
\newblock In \emph{ICASSP 2019-2019 IEEE International Conference on Acoustics,
  Speech and Signal Processing (ICASSP)}, 1030--1034. IEEE.

\bibitem[{Nahid and Kong(2019)}]{nahid2019histopathological}
Nahid, A.-A.; and Kong, Y. 2019.
\newblock Histopathological breast-image classification using concatenated
  r--g--b histogram information.
\newblock \emph{Annals of Data Science} 6(3): 513--529.

\bibitem[{Nguyen, Wang, and Nguyen(2013)}]{nguyen2013random}
Nguyen, C.; Wang, Y.; and Nguyen, H. 2013.
\newblock Random forest classifier combined with feature selection for breast
  cancer diagnosis and prognostic.
\newblock \emph{Journal of Biomedical Science and Engineering} 2013(5):
  551--560.

\bibitem[{Rebecca et~al.(2016)Rebecca, L., Siegel, MPH, Kimberly, D., Miller,
  MPH, Ahmedin, and Jemal}]{Rebecca2016Cancer}
Rebecca; L.; Siegel; MPH; Kimberly; D.; Miller; MPH; Ahmedin; and Jemal. 2016.
\newblock Cancer statistics, 2016.
\newblock \emph{Ca A Cancer Journal for Clinicians} .

\bibitem[{Roy et~al.(2019)Roy, Banik, Bhattacharjee, and
  Nasipuri}]{roy2019patch}
Roy, K.; Banik, D.; Bhattacharjee, D.; and Nasipuri, M. 2019.
\newblock Patch-based system for Classification of Breast Histology images
  using deep learning.
\newblock \emph{Computerized Medical Imaging and Graphics} 71: 90--103.

\bibitem[{Shen et~al.(2017)Shen, Zhou, Yang, Yu, Dong, Yang, Zang, and
  Tian}]{shen2017multi}
Shen, W.; Zhou, M.; Yang, F.; Yu, D.; Dong, D.; Yang, C.; Zang, Y.; and Tian,
  J. 2017.
\newblock Multi-crop convolutional neural networks for lung nodule malignancy
  suspiciousness classification.
\newblock \emph{Pattern Recognition} 61: 663--673.

\bibitem[{Song et~al.(2017)Song, Zou, Chang, and Cai}]{song2017adapting}
Song, Y.; Zou, J.~J.; Chang, H.; and Cai, W. 2017.
\newblock Adapting fisher vectors for histopathology image classification.
\newblock In \emph{2017 IEEE 14th International Symposium on Biomedical Imaging
  (ISBI 2017)}, 600--603. IEEE.

\bibitem[{Spanhol et~al.(2015)Spanhol, Oliveira, Petitjean, and
  Heutte}]{spanhol2015dataset}
Spanhol, F.~A.; Oliveira, L.~S.; Petitjean, C.; and Heutte, L. 2015.
\newblock A dataset for breast cancer histopathological image classification.
\newblock \emph{IEEE Transactions on Biomedical Engineering} 63(7): 1455--1462.

\bibitem[{Tokunaga et~al.(2019)Tokunaga, Teramoto, Yoshizawa, and
  Bise}]{tokunaga2019adaptive}
Tokunaga, H.; Teramoto, Y.; Yoshizawa, A.; and Bise, R. 2019.
\newblock Adaptive weighting multi-field-of-view CNN for semantic segmentation
  in pathology.
\newblock In \emph{Proceedings of the IEEE Conference on Computer Vision and
  Pattern Recognition}, 12597--12606.

\bibitem[{Tremblay and Borgnat(2014)}]{tremblay2014graph}
Tremblay, N.; and Borgnat, P. 2014.
\newblock Graph wavelets for multiscale community mining.
\newblock \emph{IEEE Transactions on Signal Processing} 62(20): 5227--5239.

\bibitem[{Vahadane et~al.(2015)Vahadane, Peng, Albarqouni, Baust, Steiger,
  Schlitter, Sethi, Esposito, and Navab}]{vahadane2015structure}
Vahadane, A.; Peng, T.; Albarqouni, S.; Baust, M.; Steiger, K.; Schlitter,
  A.~M.; Sethi, A.; Esposito, I.; and Navab, N. 2015.
\newblock Structure-preserved color normalization for histological images.
\newblock In \emph{2015 IEEE 12th International Symposium on Biomedical Imaging
  (ISBI)}, 1012--1015. IEEE.

\bibitem[{Veeling et~al.(2018)Veeling, Linmans, Winkens, Cohen, and
  Welling}]{veeling2018rotation}
Veeling, B.~S.; Linmans, J.; Winkens, J.; Cohen, T.; and Welling, M. 2018.
\newblock Rotation equivariant CNNs for digital pathology.
\newblock In \emph{International Conference on Medical image computing and
  computer-assisted intervention}, 210--218. Springer.

\bibitem[{Wang et~al.(2020)Wang, Chen, Lu, Baras, and Mahmood}]{wang2020weakly}
Wang, J.; Chen, R.~J.; Lu, M.~Y.; Baras, A.; and Mahmood, F. 2020.
\newblock Weakly supervised prostate tma classification via graph convolutional
  networks.
\newblock In \emph{2020 IEEE 17th International Symposium on Biomedical Imaging
  (ISBI)}, 239--243. IEEE.

\bibitem[{Wang et~al.(2018)Wang, Dong, Dai, Rosario, and
  Xing}]{wang2018classification}
Wang, Z.; Dong, N.; Dai, W.; Rosario, S.~D.; and Xing, E.~P. 2018.
\newblock Classification of breast cancer histopathological images using
  convolutional neural networks with hierarchical loss and global pooling.
\newblock In \emph{International Conference Image Analysis and Recognition},
  745--753. Springer.

\bibitem[{Xu et~al.(2019)Xu, Shen, Cao, Qiu, and Cheng}]{xu2019graph}
Xu, B.; Shen, H.; Cao, Q.; Qiu, Y.; and Cheng, X. 2019.
\newblock Graph wavelet neural network.
\newblock \emph{arXiv preprint arXiv:1904.07785} .

\bibitem[{Yan et~al.(2020)Yan, Ren, Wang, Wang, Zhang, Liu, Rao, Zheng, and
  Zhang}]{yan2020breast}
Yan, R.; Ren, F.; Wang, Z.; Wang, L.; Zhang, T.; Liu, Y.; Rao, X.; Zheng, C.;
  and Zhang, F. 2020.
\newblock Breast cancer histopathological image classification using a hybrid
  deep neural network.
\newblock \emph{Methods} 173: 52--60.

\bibitem[{Yao et~al.(2019)Yao, Zhang, Zhou, and Liu}]{yao2019parallel}
Yao, H.; Zhang, X.; Zhou, X.; and Liu, S. 2019.
\newblock Parallel structure deep neural network using cnn and rnn with an
  attention mechanism for breast cancer histology image classification.
\newblock \emph{Cancers} 11(12): 1901.

\bibitem[{Zhou et~al.(2018)Zhou, Hang, Liu, and Yuan}]{zhou2018integrating}
Zhou, F.; Hang, R.; Liu, Q.; and Yuan, X. 2018.
\newblock Integrating convolutional neural network and gated recurrent unit for
  hyperspectral image spectral-spatial classification.
\newblock In \emph{Chinese Conference on Pattern Recognition and Computer
  Vision (PRCV)}, 409--420. Springer.

\bibitem[{Zhou et~al.(2019)Zhou, Graham, Alemi~Koohbanani, Shaban, Heng, and
  Rajpoot}]{zhou2019cgc}
Zhou, Y.; Graham, S.; Alemi~Koohbanani, N.; Shaban, M.; Heng, P.-A.; and
  Rajpoot, N. 2019.
\newblock Cgc-net: Cell graph convolutional network for grading of colorectal
  cancer histology images.
\newblock In \emph{Proceedings of the IEEE International Conference on Computer
  Vision Workshops}, 0--0.

\end{thebibliography}
\end{document}